\documentclass[runningheads]{llncs}
\usepackage{graphicx}
\usepackage{amsmath,amssymb} 
\usepackage{color}
\usepackage[width=122mm,left=12mm,paperwidth=146mm,height=193mm,top=12mm,paperheight=217mm]{geometry}
\begin{document}
\pagestyle{headings}
\mainmatter

\title{A generalised feature for low level vision} 

\author{Dr D.A.Sinclair, Dr C.P.Town}
\institute{david@imense.com, Imense Ltd. \\cpt23@cam.ac.uk, Department of Computer Science, \\University of Cambridge.
\\3rd Feb 2021}

\maketitle

\begin{abstract}
  This papers presents a novel quantised transform (the Sinclair-Town or ST transform for short)
  that subsumes the rolls of both edge-detector, MSER style region detector and corner detector.
  The transform is similar to the \( unsharp \) transform but the difference from the local mean is 
  quantised to 3 values (dark-neutral-light). The transform naturally leads to the definition of an
  appropriate local scale. A range of methods for extracting shape features form the transformed image are presented.
  The generalised feature provides a robust basis for establishing correspondence between images.
  The transform readily admits more complicated kernel behaviour including multi-scale and asymmetric
  elements to prefer shorter scale or oriented local features.

\keywords{Stereo matching, visual correspondence, quantised transform, shape features}
\end{abstract}

\section{Introduction}

There exist a range of methods for extracting primary low-level features from images.
This include edge, corner and blob detection. Edge detection methods vary from convolution
based measurement of local image gradient \cite{Canny86} to morphology based
methods \cite{Smith97}. Numerous corner and blob detectors now exist with noteworthy
methods being SIFT \cite{Lowe04} (local gradient based)  and MSER \cite{Matas02robustwide} (area topology based).
Low level features are typically used to establish correspondence
between images or as indexing terms in object recognition or navigation.
The low level feature detector presented here is defined to be the 3-value-quantised difference between
a pixel value and an estimate of the local image mean. We have christened this the Sinclair-Town or
ST-transform for short. In spirit this falls between an edge detector and the MSER blob detector.
The transform is efficient to compute as rectangular areas can be used to compute the local mean
via the cumulative sum image. The range of corner and interest point detectors is sufficiently large
that comment on such is left to survey papers and this paper will focus on exposition of the ST transform,
however useful papers do include \cite{Dahl_findingthe}, \cite{FTombari08} \cite{DBLP:conf/bmvc/RichardsFJ92}.
Section \ref{section:1} gives details of the transform and examples of its output.
Section \ref{section:2} show how the output of the ST-transform can be used to define various extended
features. 
Section \ref{section:3} demonstrates the use of ST-transform features in image matching and section \ref{section:4}
offers wise and considered conclusions.

\section{The Sinclair-Town (ST) transform.}
\label{section:1}

We define the ST (Sinclair-Town) transform of an image I to be:

\begin{equation} 
if( (I - m(I,d)) > k1 ) ST(I) = 1,\\
if( (I - m(I,d)) < -k2 ) ST(I) = -1\\
else ST(I) = 0.
\end{equation}

Where m(I,d) is the local mean of I computed over a region of side length 2*d+1 centred on the pixel. k1 and k2 are
typically set to 4 for stereo matching and d to 12. The local mean is efficiently estimated using a cumulative sum
of image brightness.

Matlab code for the single scale ST transform.
\noindent
\begin{verbatim}
function trx = ST(im, d,k1,k2)
% Sinclair-Town transform
%
% trx = ST(double(im), 12, 4, 4);
%

q = d*2+1;
D = d+1;
[nr,nc] = size(im);
s = cumsum(cumsum(im,2),1); 
br = nr-d-1;
bc = nc-d-1;
b0 = D+1;
trx = zeros(nr,nc);
nx = q*q;
for r=b0:br
    for c=b0:bc
        % the Sinclair-Town transform is the trinarised difference from the local mean.
        v= (s(r-D,c-D)-s(r-D,c+d)-s(r+d,c-D)+s(r+d,c+d))/nx;
        k = im(r,c)-v;
        if k > k1
            trx(r,c) = 1;
        elseif k < -k2
            trx(r,c) = -1;
        end
    end
end
\end{verbatim}

\begin{figure}
\centering$
\begin{array}{ll}
\includegraphics[width=.48\textwidth]{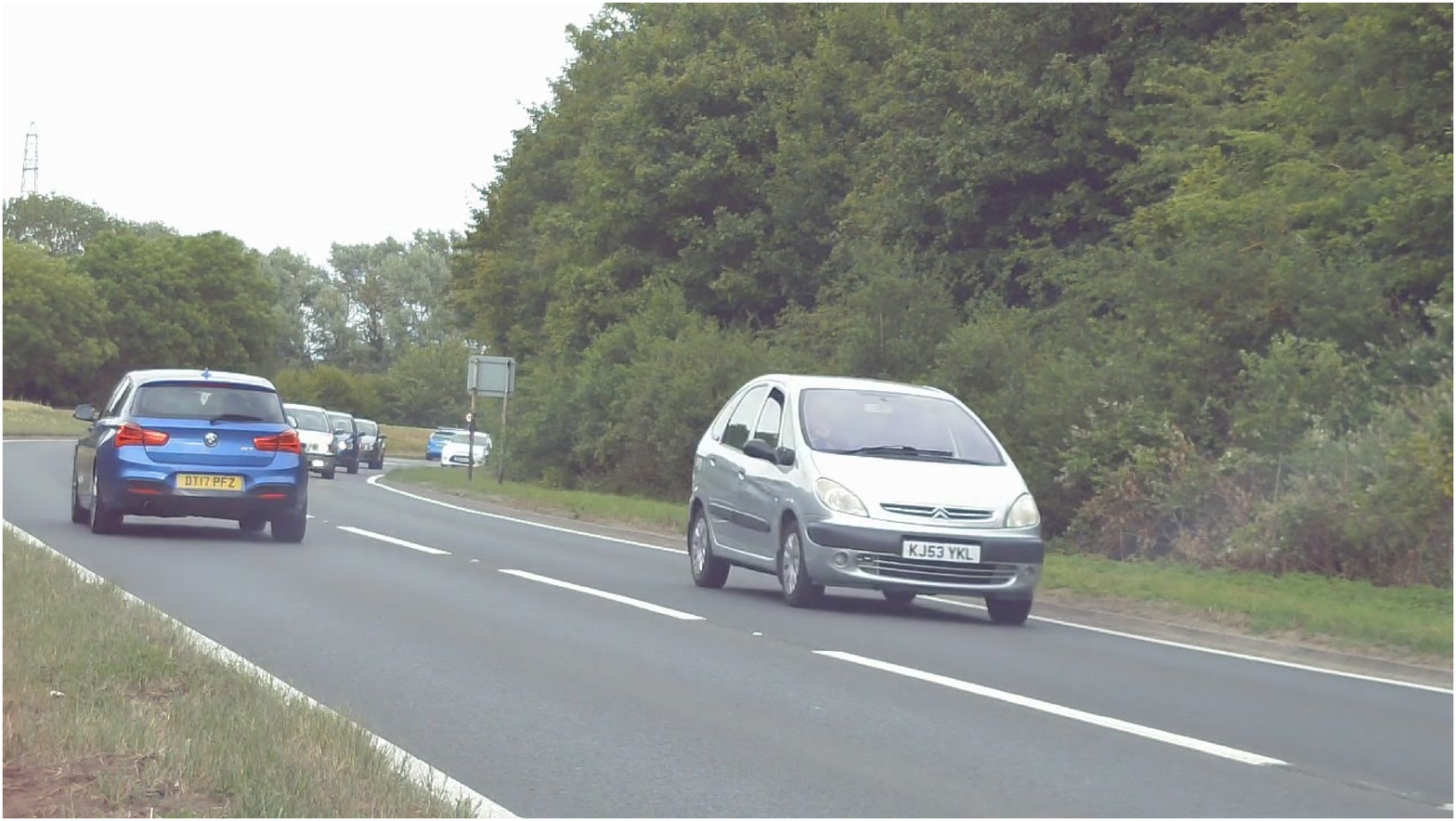}&
\includegraphics[width=.48\textwidth]{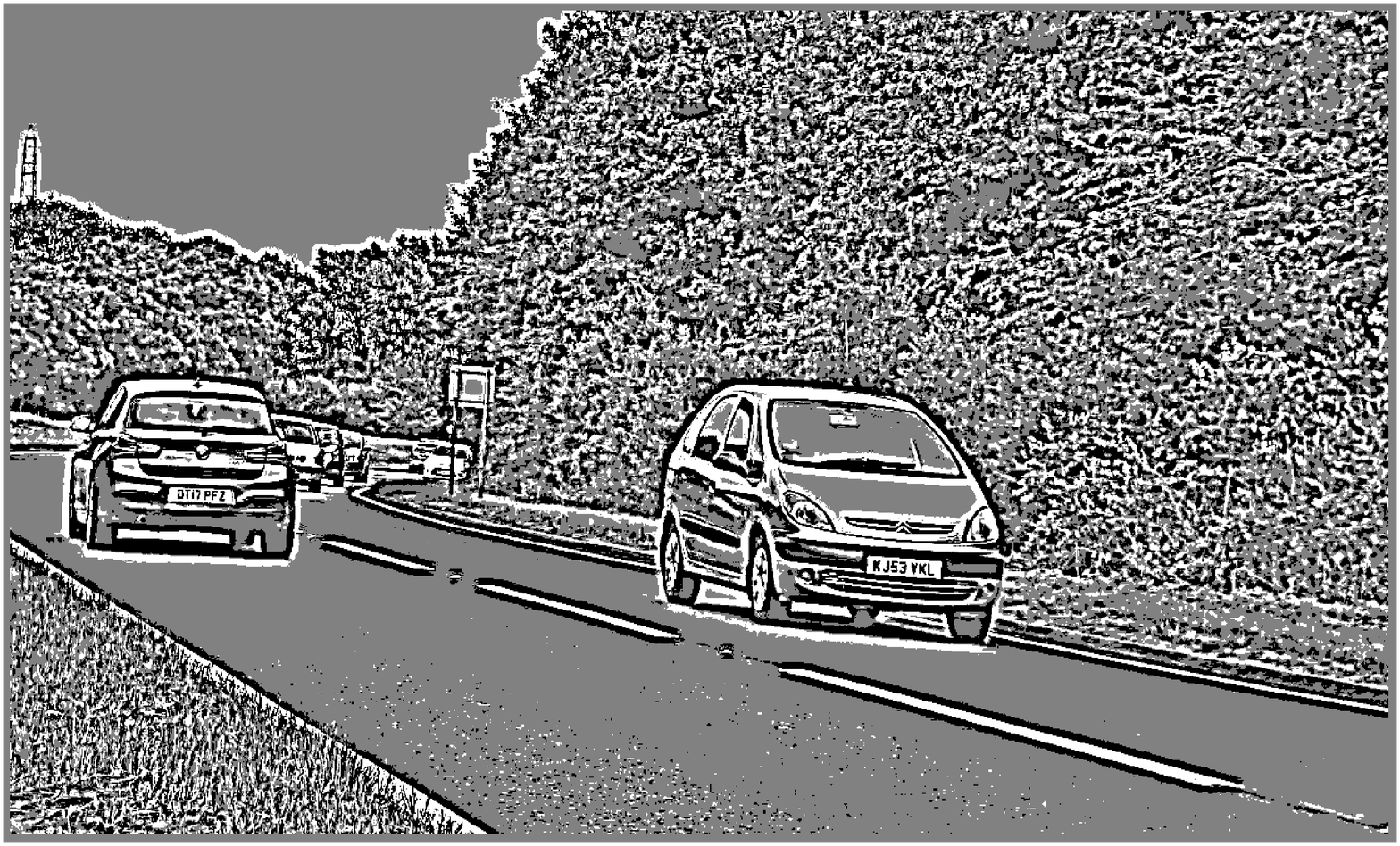} \\
a. & b. \\
\includegraphics[width=.48\textwidth]{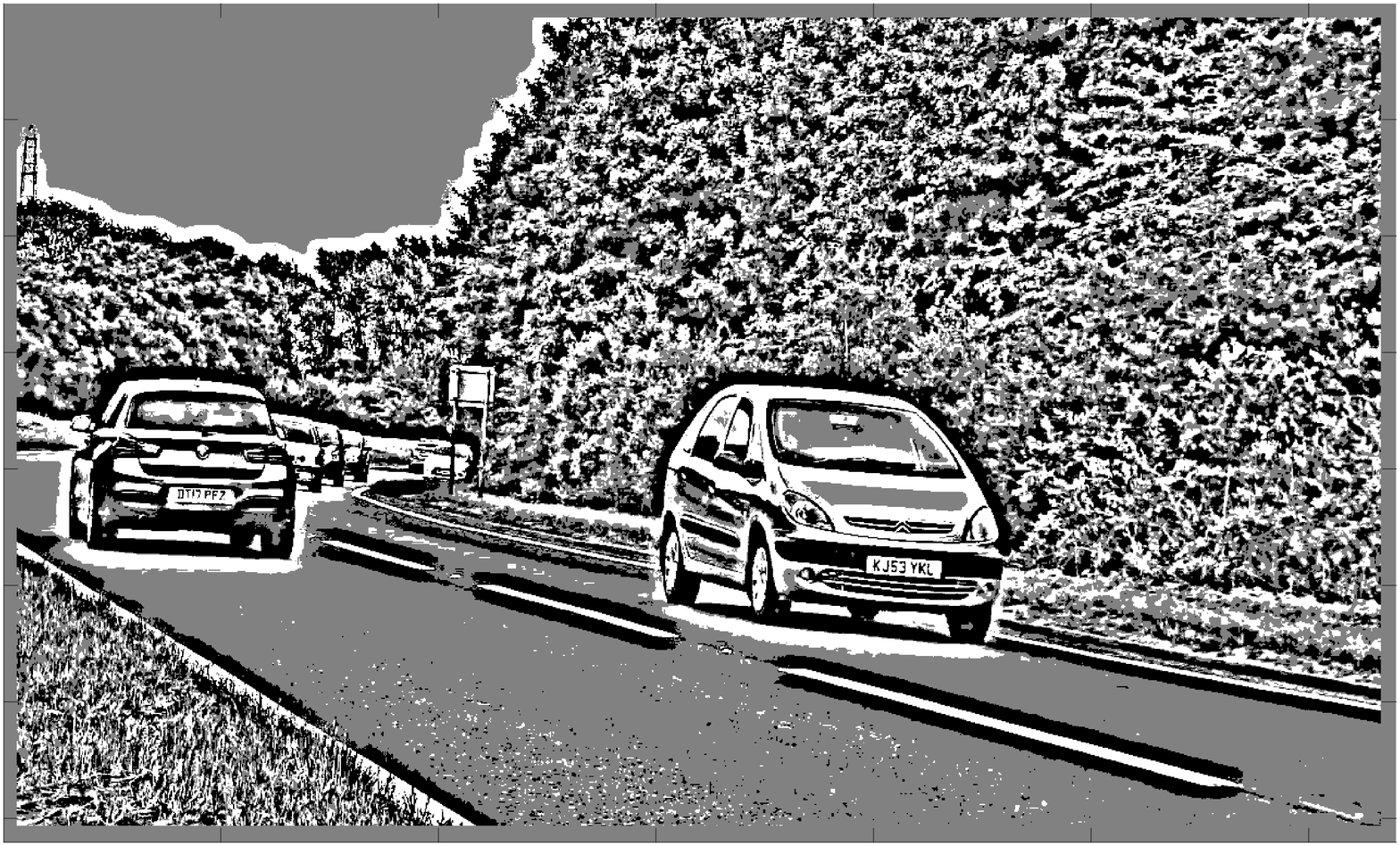} \\
c. \\
\end{array}$
\caption{
\emph{a.} Image from a super interesting video sequence of passing cars. \emph{b.} ST-transform with \{d=6\}.
\emph{c.} ST-transform with \{d=12\}.   
}
\label{fig:ST_1}
\end{figure}

Figure~\ref{fig:ST_1} shows an image of some cars moving on a road and the associated ST transforms for d=6,12,18 k=4.
Increasing d gives a wider band round dark-light edges and increases tolerance to change in shape under correlation. 

\begin{figure}
  \centering$
  \begin{array}{ll}
    \includegraphics[width=.48\textwidth]{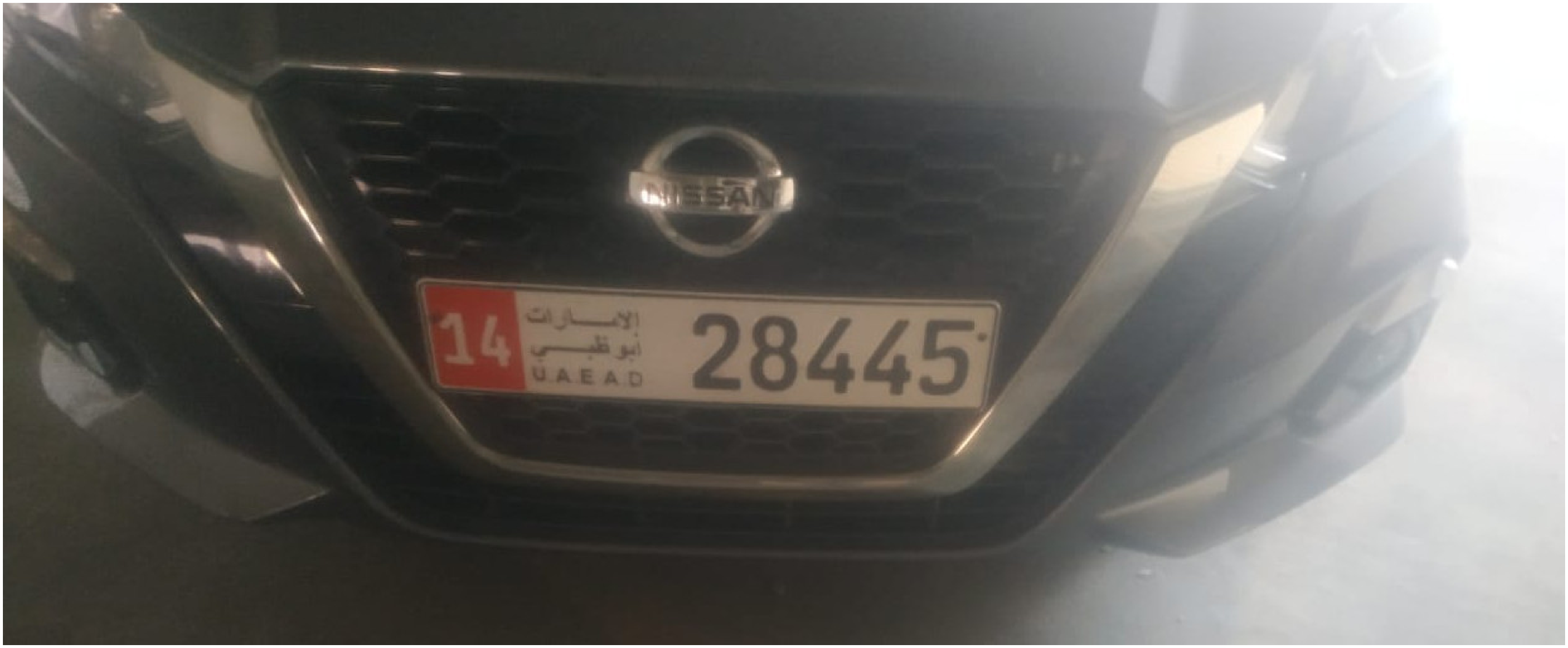}&
      \includegraphics[width=.48\textwidth]{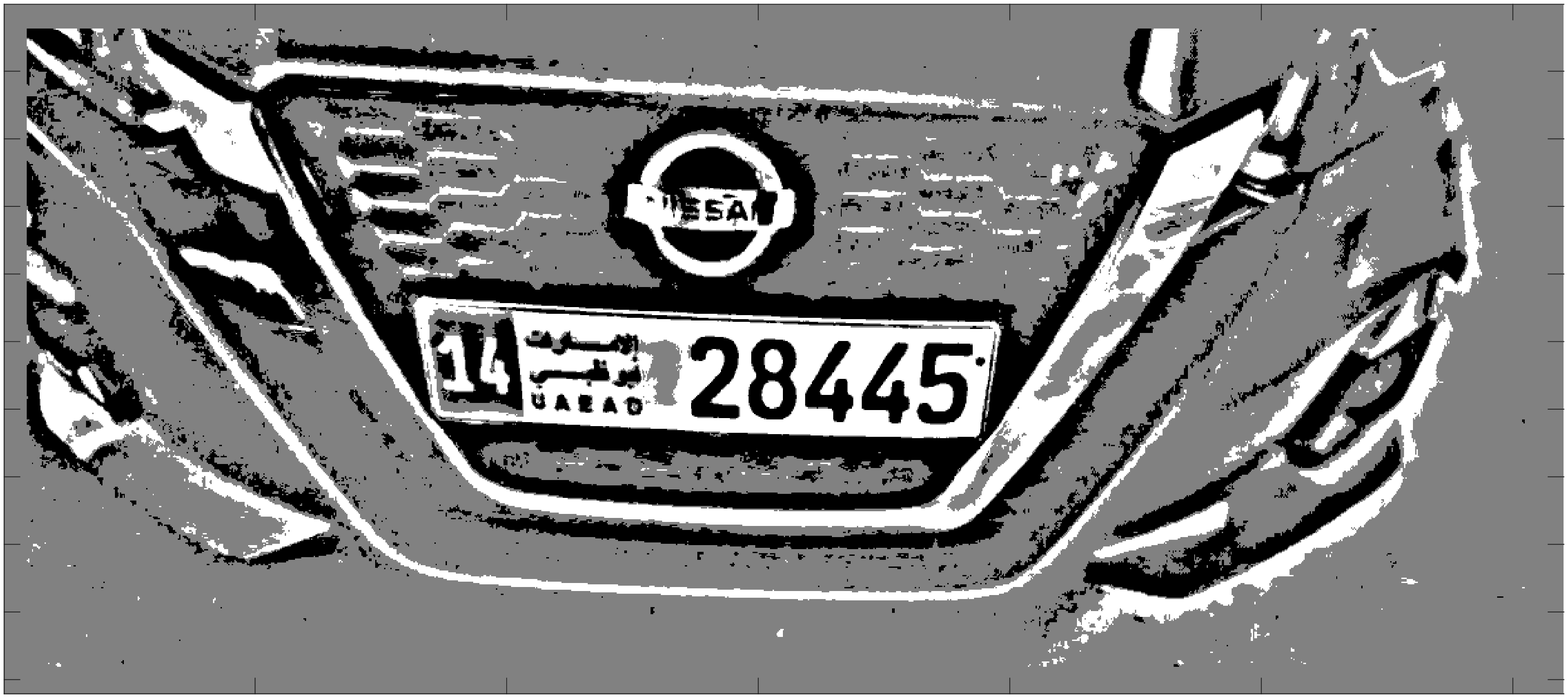}\\
        a. & b.\\
        
        \includegraphics[width=.48\textwidth]{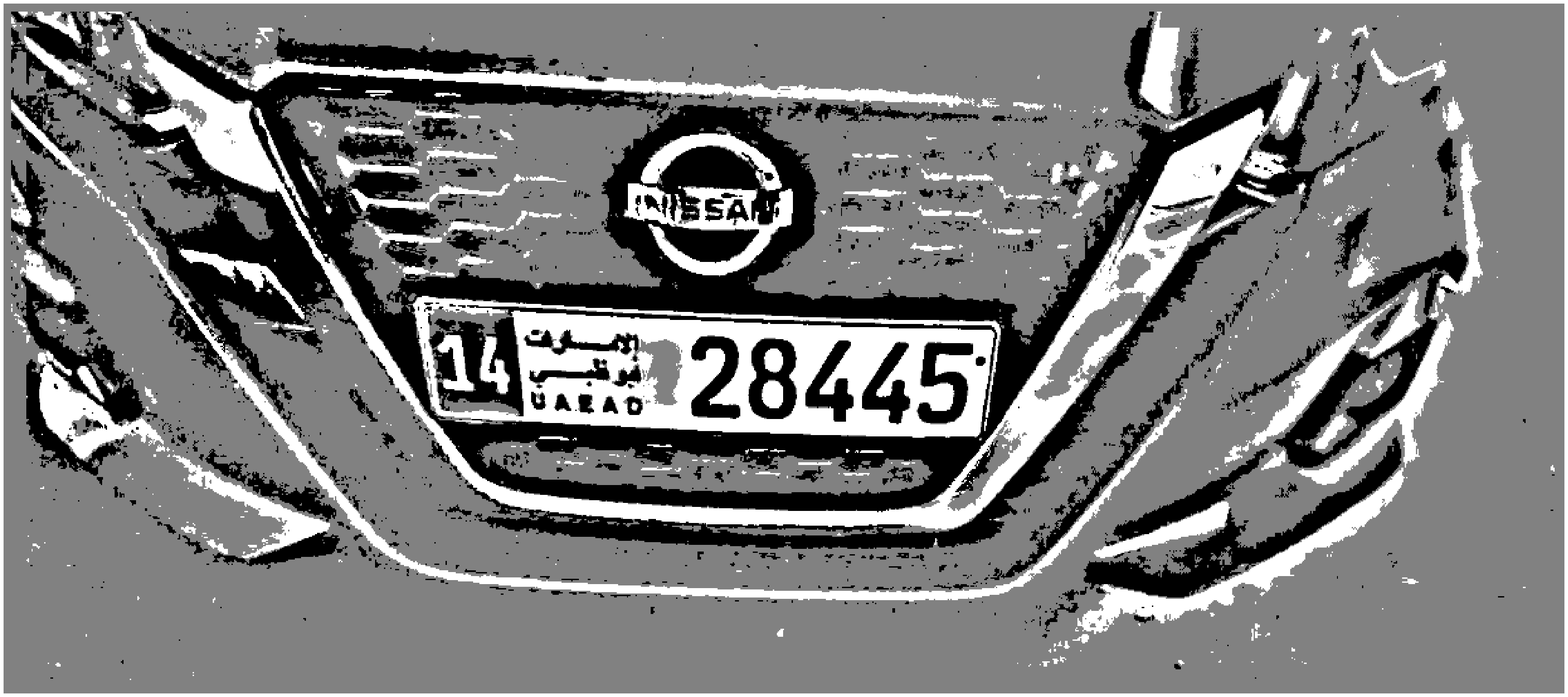}\\
          c.\\
          
  \end{array}$

  \caption{
    \emph{  a.} Super interesting image of a car license plate. \emph{b.} single scale ST-transform.
    \emph{c.} multi-scale ST-transform.
    Adding a second scale to the TS transform allows fine detail to dominate over strong longer scale brightness differences.
  }
  \label{fig:ST_2}
\end{figure}

The ST-transform can be extended to cope with detail on multiple scales by introducing more than one
estimate of local mean with differing sizes of domains of support (fig.~\ref{fig:ST_2}).
Sensitivity to local edges can also
be increased through the inclusion of local asymmetric mean estimates.

\section{Feature extraction from the ST transform.}
\label{section:2}

\begin{figure}
\centering$
  \begin{array}{l}
    \includegraphics[width=.9\textwidth]{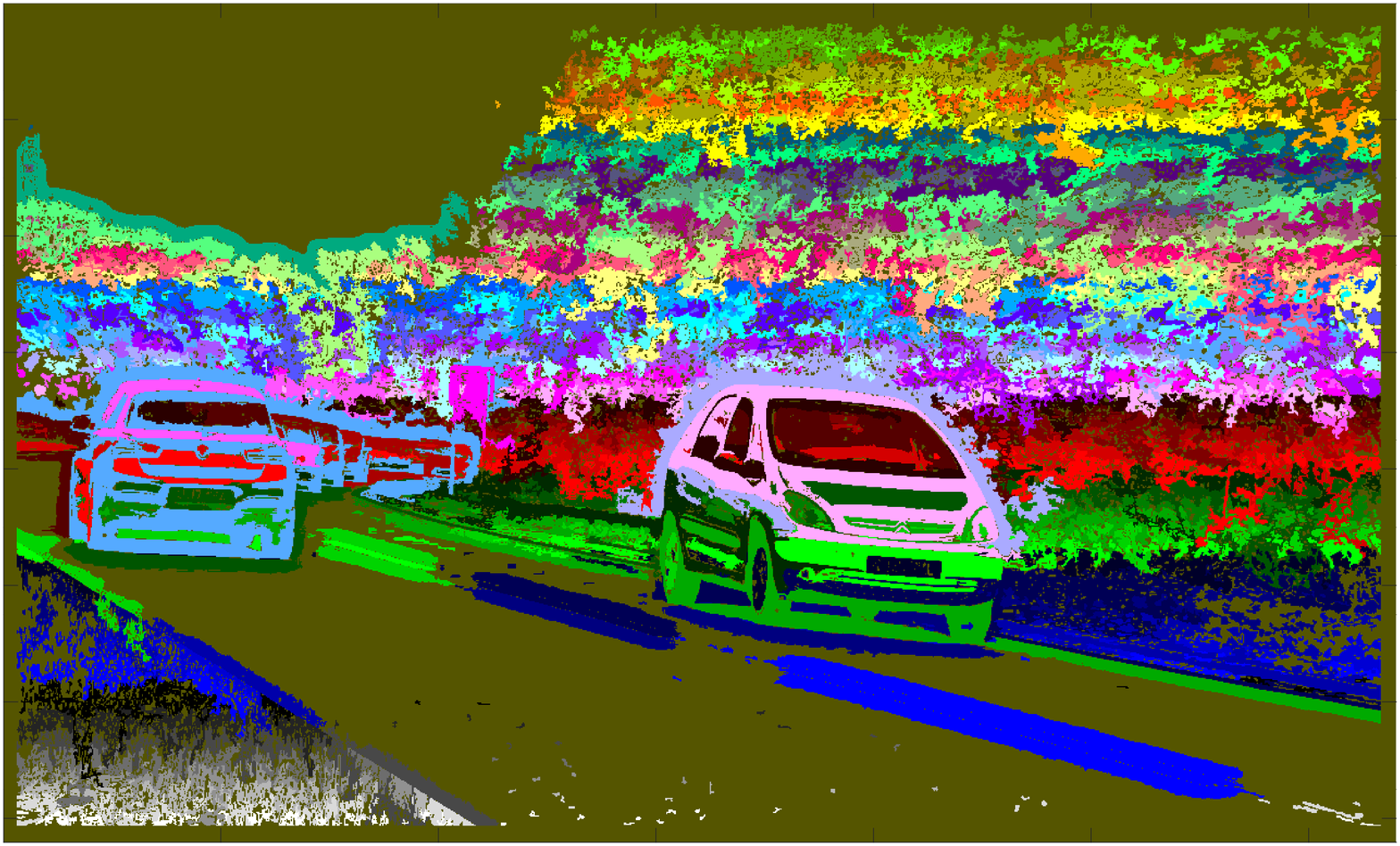}\\
      \includegraphics[width=.9\textwidth]{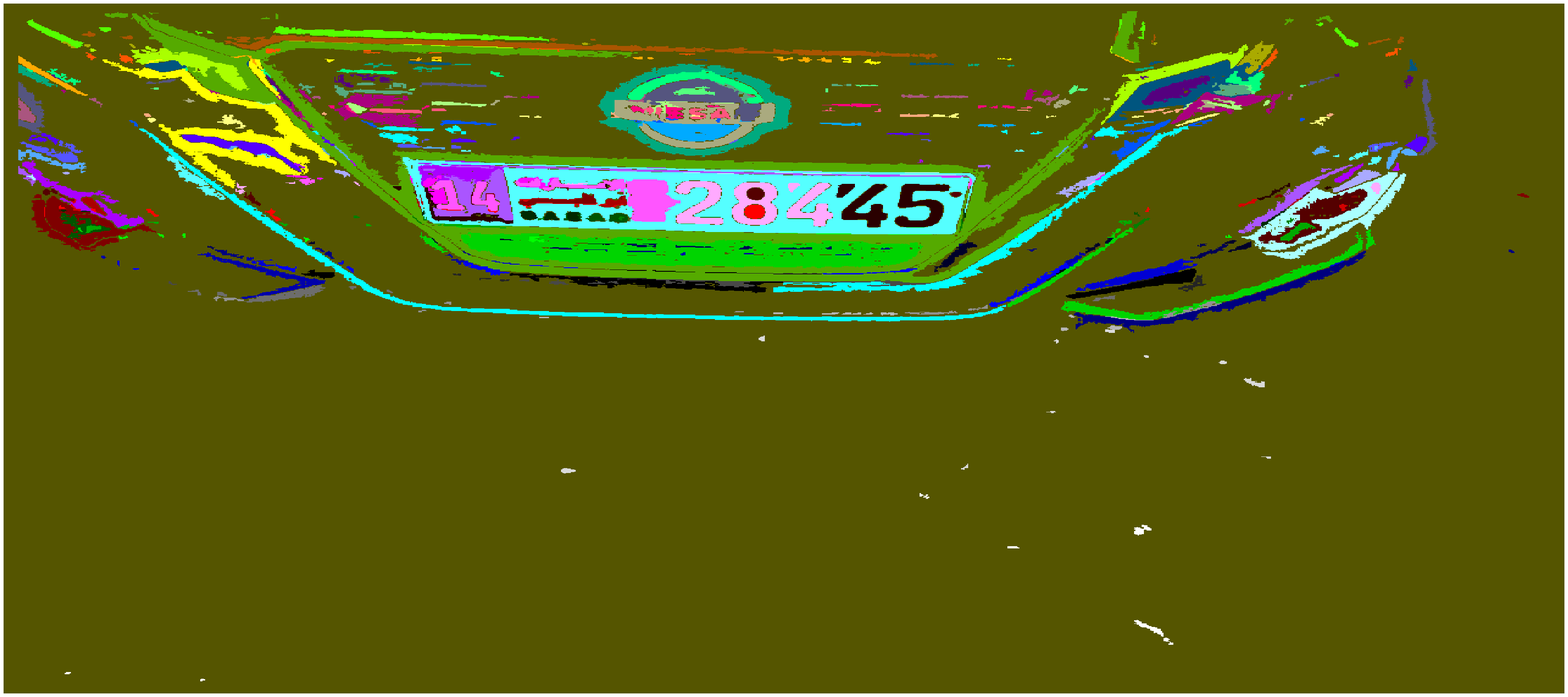}\\

  \end{array}$

\caption{
Mutlab artificially coloured regions of constant quantisation value form figures \ref{fig:ST_1}
and \ref{fig:ST_2}. 
}
\label{fig:ST_3}
\end{figure}

The ST transform produces a 3 valued quantised result representing areas that are either
1 (brighter than the local mean by a threshold), -1 (darker than the local mean by a threshold) or
0 (similar to the local mean).
The simplest feature to extract from the quantised image is areas of constant quantisation.
Figure ~\ref{fig:ST_3} shows regions of constant quantisation for the images in figure~\ref{fig:ST_1}
and ~\ref{fig:ST_2}.

These area features are faster to compute than MSER regions or multiscale SIFT features
and are generally stable to mild change in viewpoint.

\subsection{Edge and corner features from the ST transform}

\begin{figure}
  \centering
  \includegraphics[width=.9\textwidth]{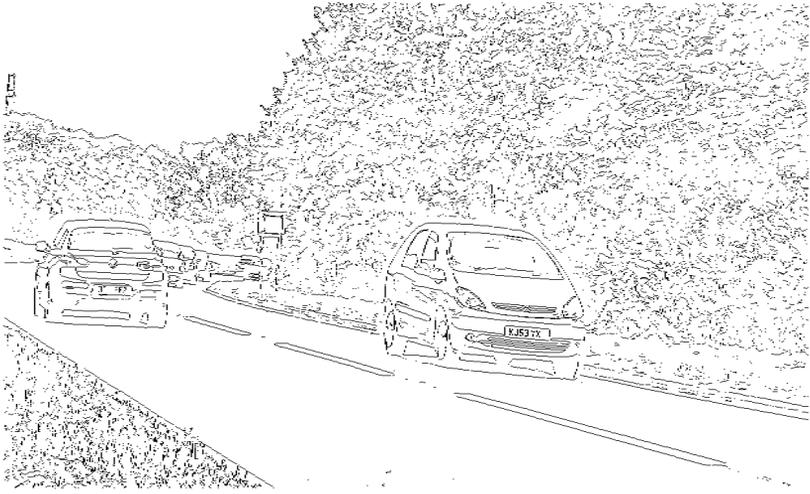}
  \caption{
    Dark ST edges of figure \ref{fig:ST_1}. 
  }
  \label{fig:ST_4}
\end{figure}

The ST-transform may be used as an edge detector by labelling for example pixels on the boundary
of dark regions that are adjacent or close to adjacent to light pixels as edges: figure  ~\ref{fig:ST_4}.
Equally a gradient threshold (original image gradient across the boundary) could be applied to boundary
pixels of dark regions.

\begin{figure}
\centering
  \includegraphics[width=.9\textwidth]{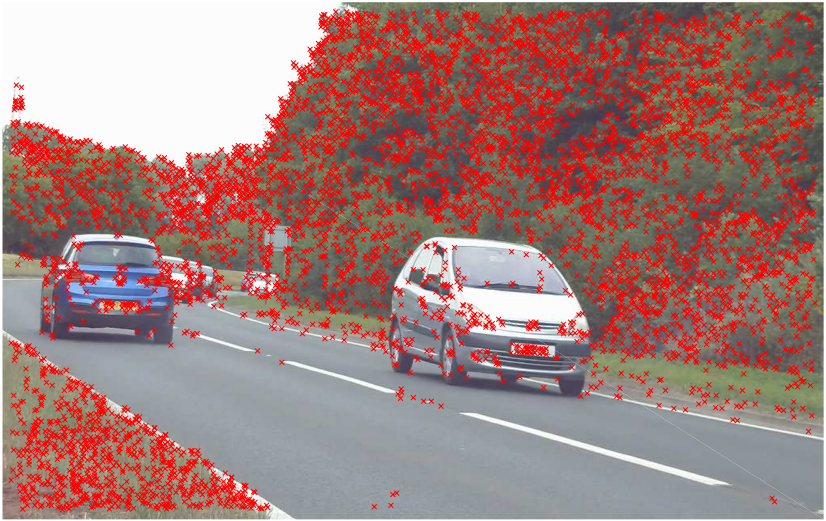}

  \caption{
    Maxima of curvature of dark ST-edges from \ref{fig:ST_1} overlayed. 
  }
  \label{fig:ST_5}
\end{figure}

Corners may be defined as maxima of curvature of dark (or light edges): figure  ~\ref{fig:ST_5}.
A length scale is required for curvature estimation, in this example +/- 4 pixels along the edge chain was used.

\section{Matching with the ST-transform.}
\label{section:3}

The ST-transform of an extended area provides a stable basis for establishing correspondence
between areas of an image and other images in a sequence or from a stereo pair.
A range of metrics are possible for comparing either the intensity values or the ST-quantised values
of two image patches. For the purposes of this paper we demonstrate a sum of absolute differences metric.

\begin{figure}
\centering$
\begin{array}{ll}
\includegraphics[width=.48\textwidth]{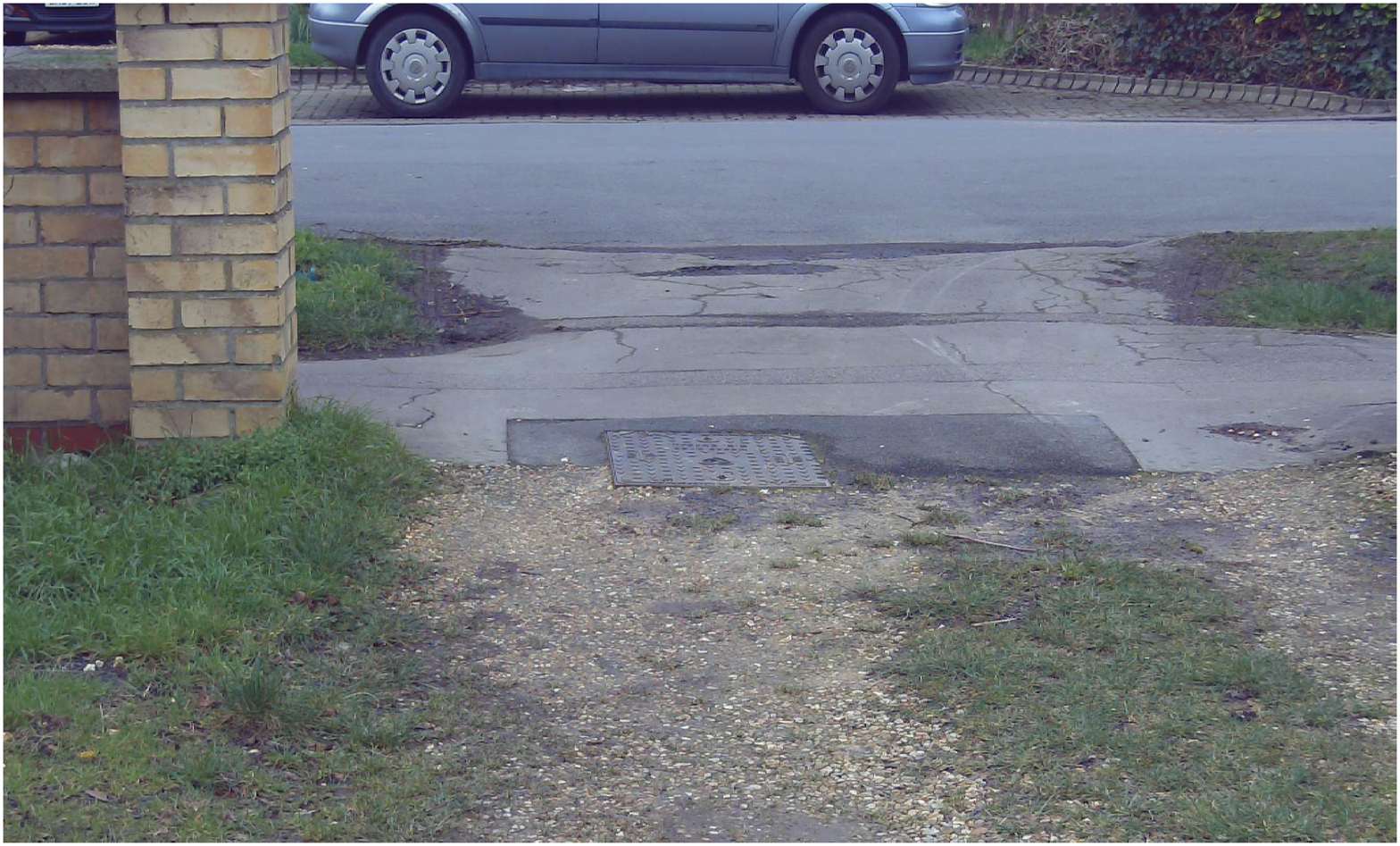}&
\includegraphics[width=.48\textwidth]{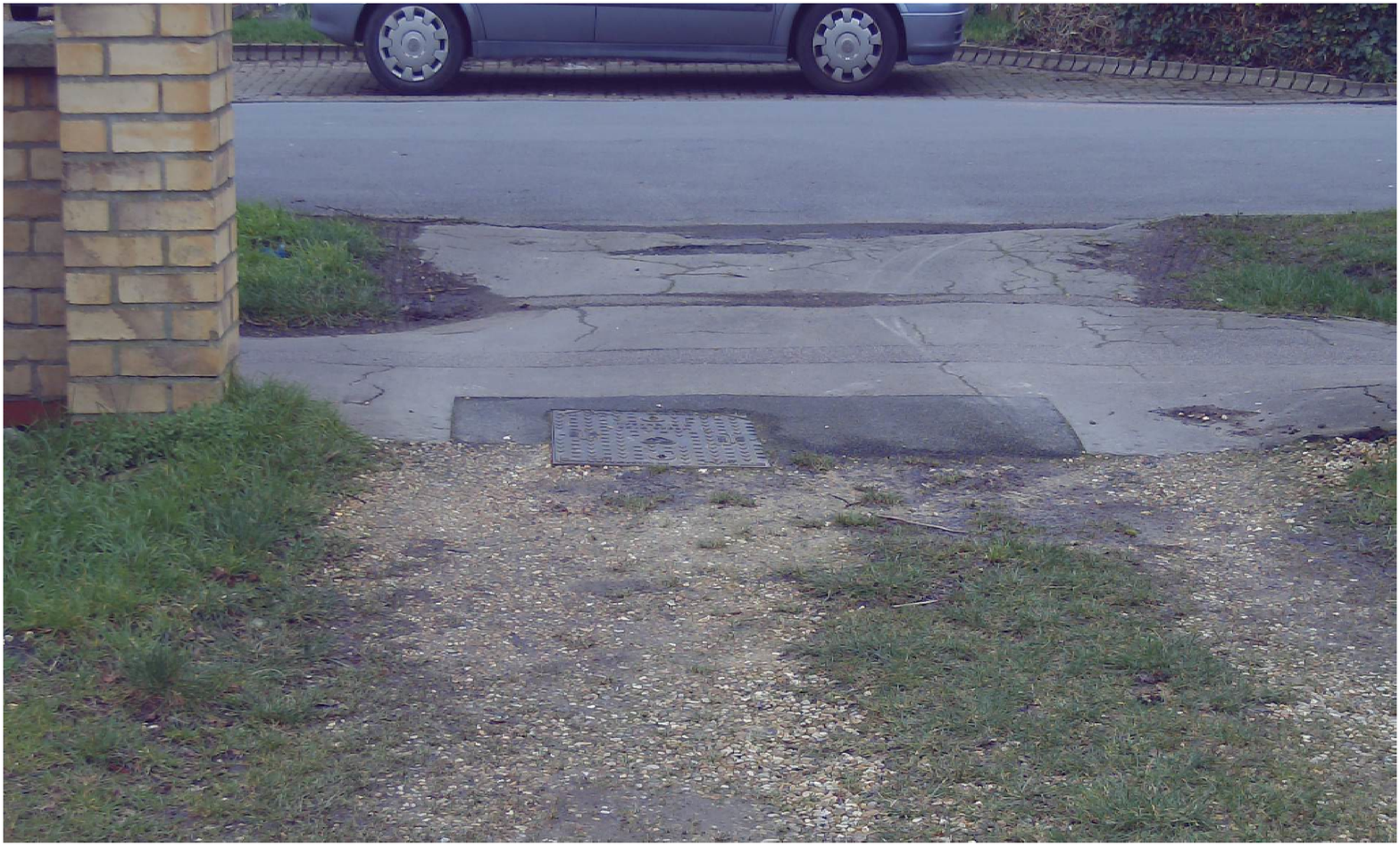} \\
a. & b. \\
\includegraphics[width=.48\textwidth]{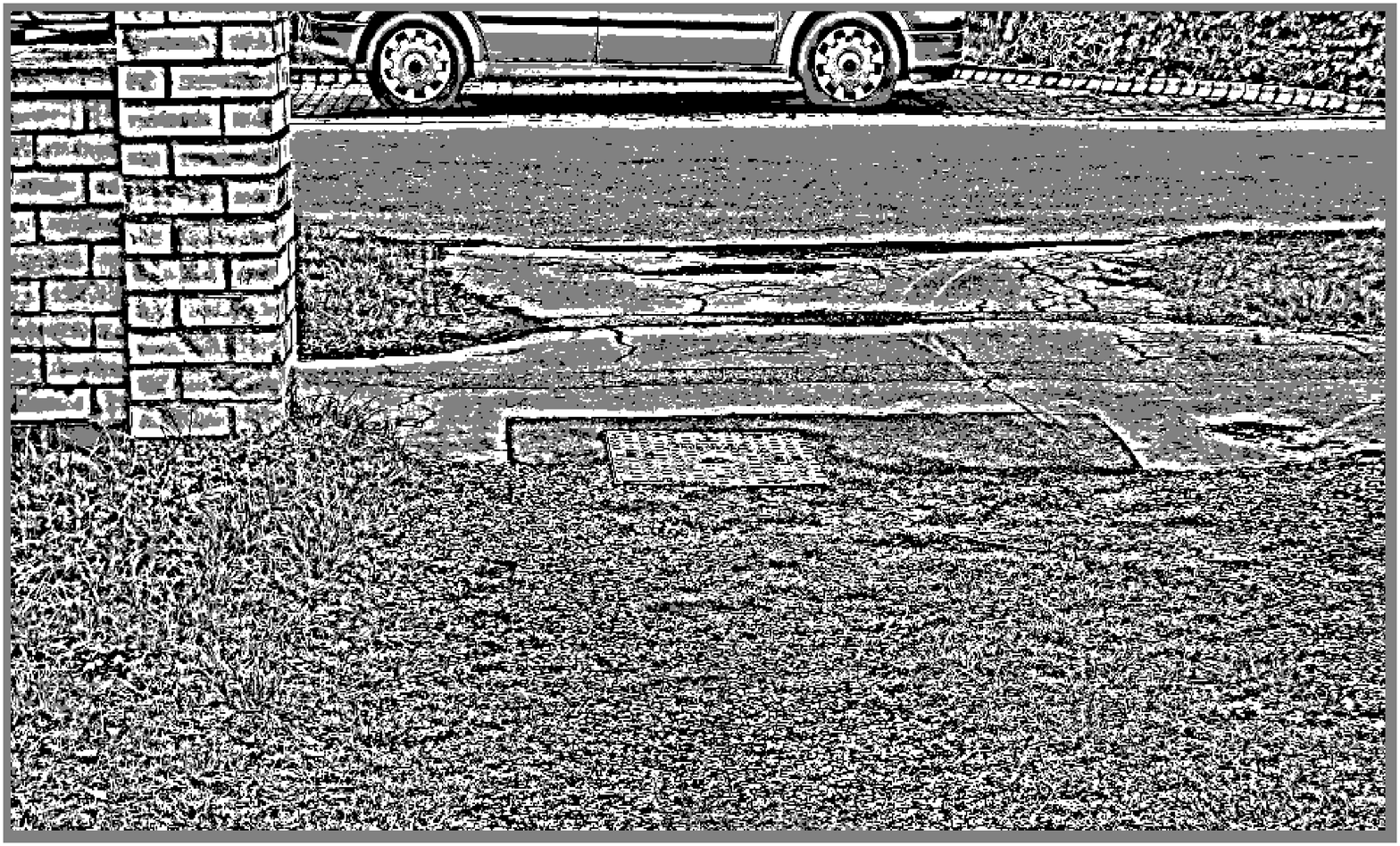}&
\includegraphics[width=.48\textwidth]{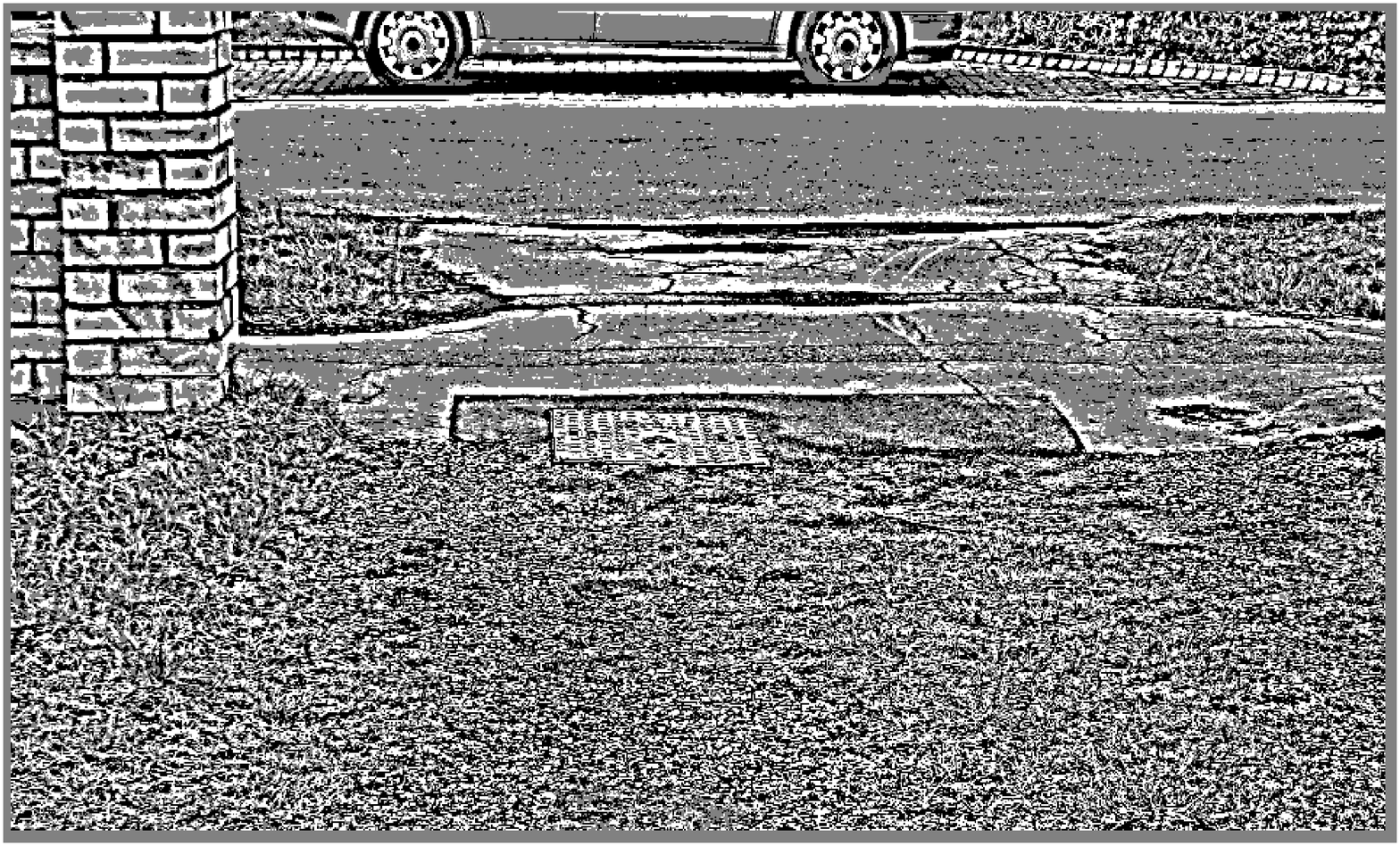}\\
c. & d. \\
\end{array}$
\caption{
\emph{a.} Image from a plump, ripe stereo pair. \emph{b.} other image form the pair \{d=6\}.
\emph{c.} ST-transform of \emph{a} with \{d=12\} \emph{d.} ST-transform of \emph{b} with \{d=12\} .   
}
\label{fig:ST_6}
\end{figure}

Figure ~\ref{fig:ST_6} shows a pair of images from a stereo camera a rig and their ST-transformed versions
with a mean area patch dimension of 25x25 pixels.

\begin{figure}
\centering$
\begin{array}{l}
\includegraphics[width=.75\textwidth]{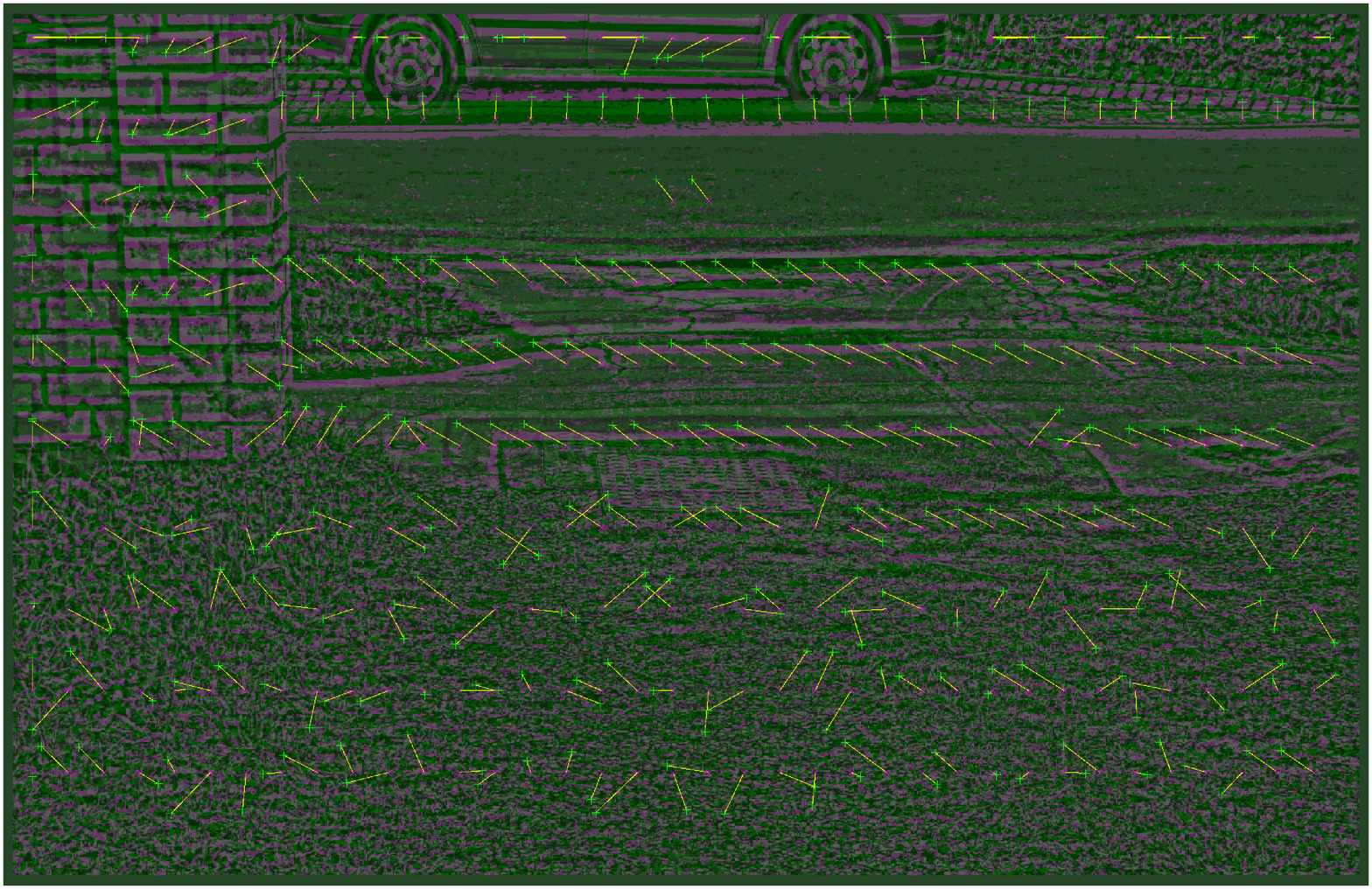}\\
\includegraphics[width=.9\textwidth]{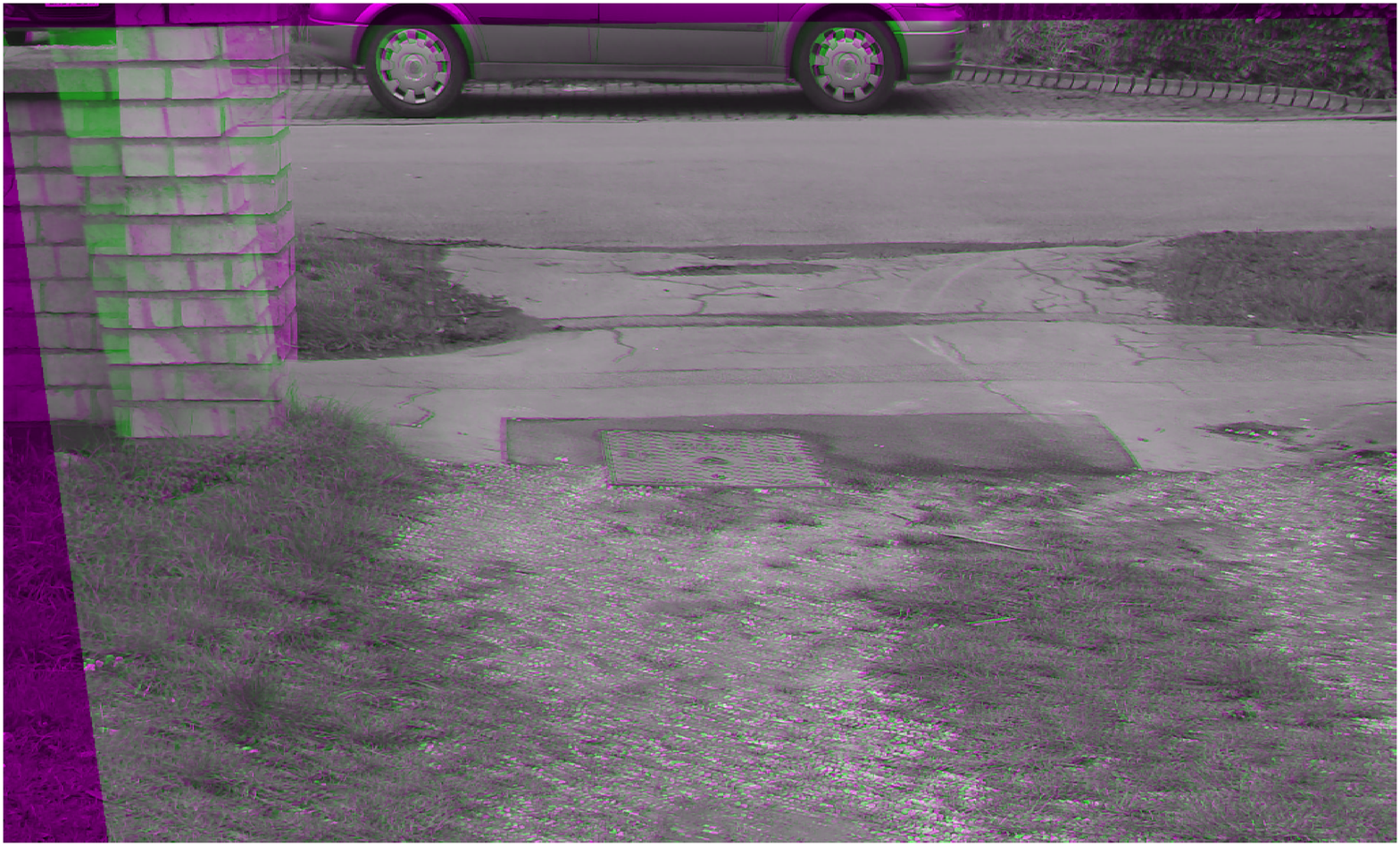} \\
\end{array}$
\caption{
Block matches created by the Matlab matching code, and a projectively rectified overlay of the image pair. 
}
\label{fig:ST_7}
\end{figure}

Figure ~\ref{fig:ST_7} shows a block matches between the stereo pair in figure ~\ref{fig:ST_6} together
with the images brought into registration via the projective transform of the plane of the road, estimated
from the matches \cite{Sinclair92}.

Figure ~\ref{fig:ST_7} shows a block matches between the amply, gorgeous stereo pair in figure ~\ref{fig:ST_6} together
with the images brought into registration via the projective transform of the plane of the road,
estimated from the matches.

The projective distortion of the driveway increases towards the bottom of the field of view and matching is degraded.
In reality it would be best to include an interactive projective rectification step during the matching process or use
afinely normalised indexing features.

\begin{figure}
\centering$
\begin{array}{ll}
\includegraphics[width=.48\textwidth]{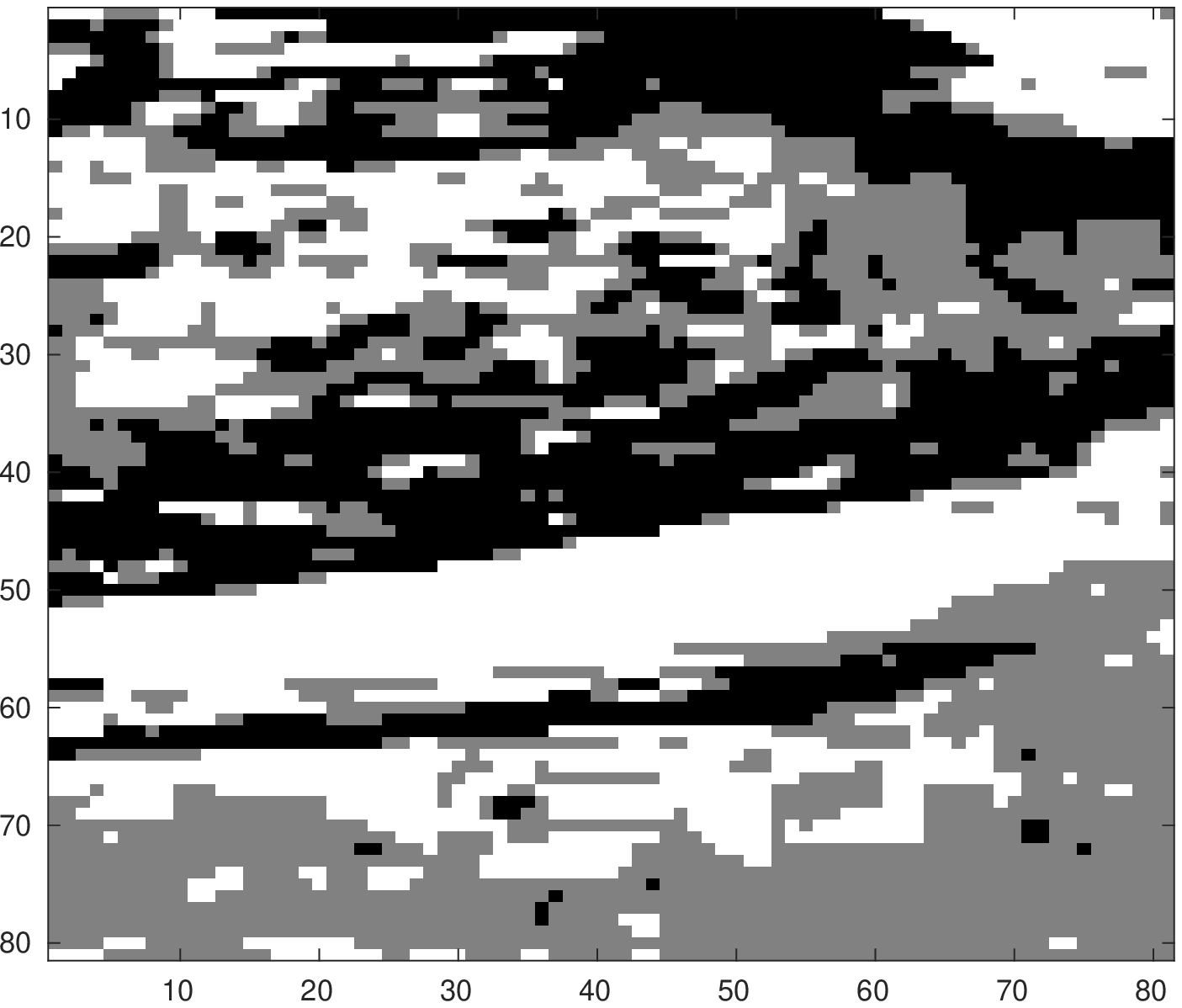}&
\includegraphics[width=.48\textwidth]{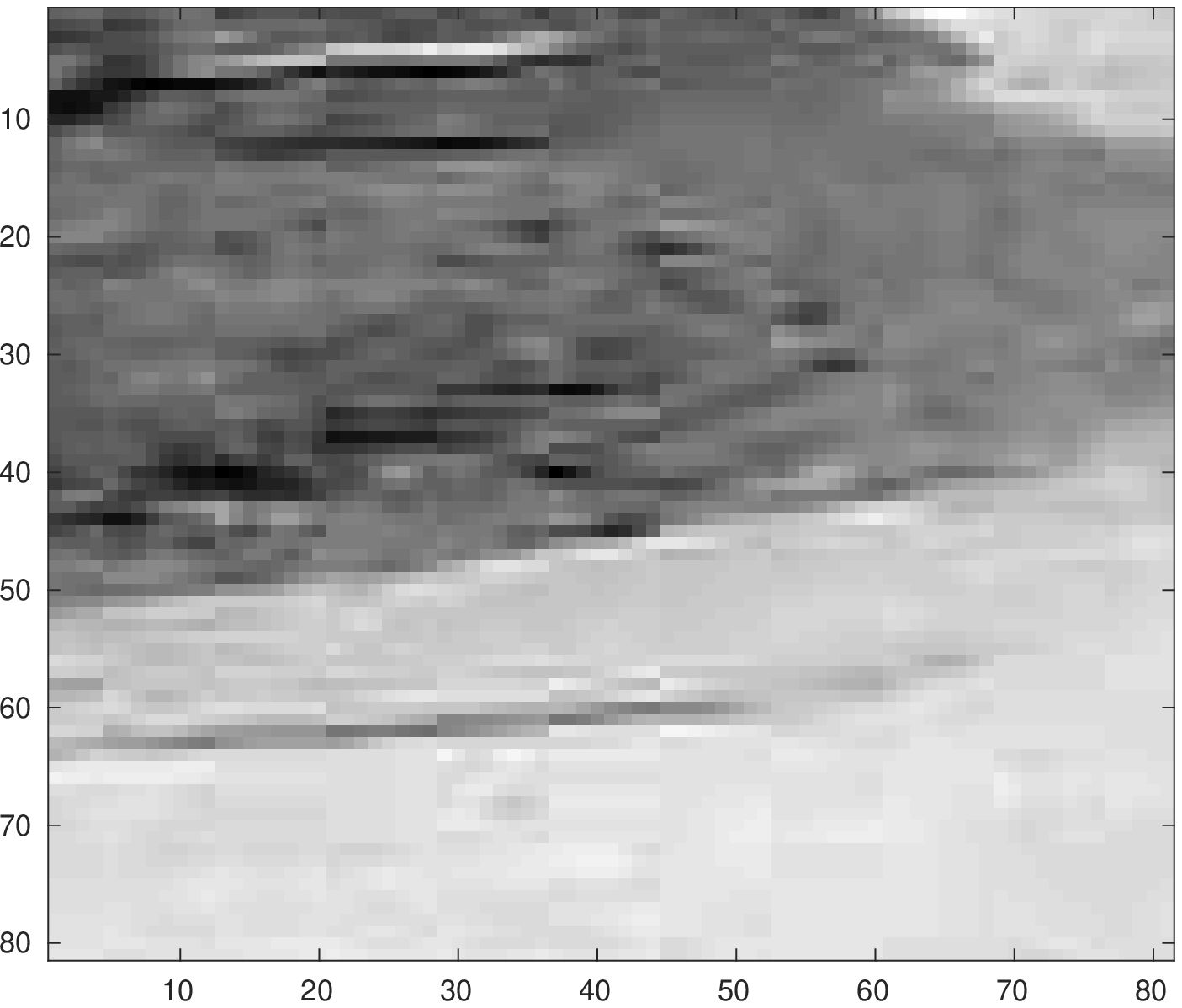} \\
a. & b. \\
\includegraphics[width=.48\textwidth]{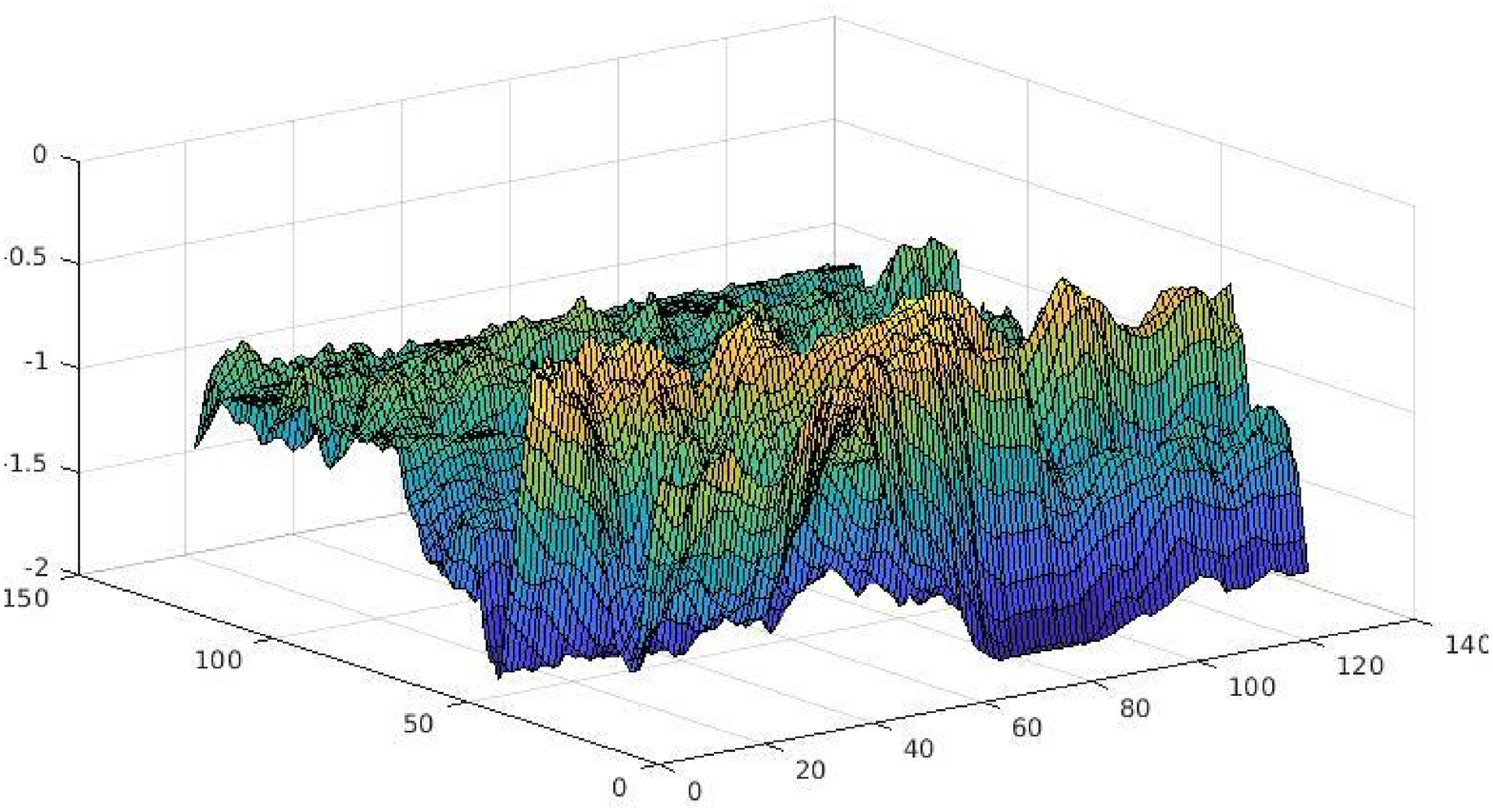}&
\includegraphics[width=.48\textwidth]{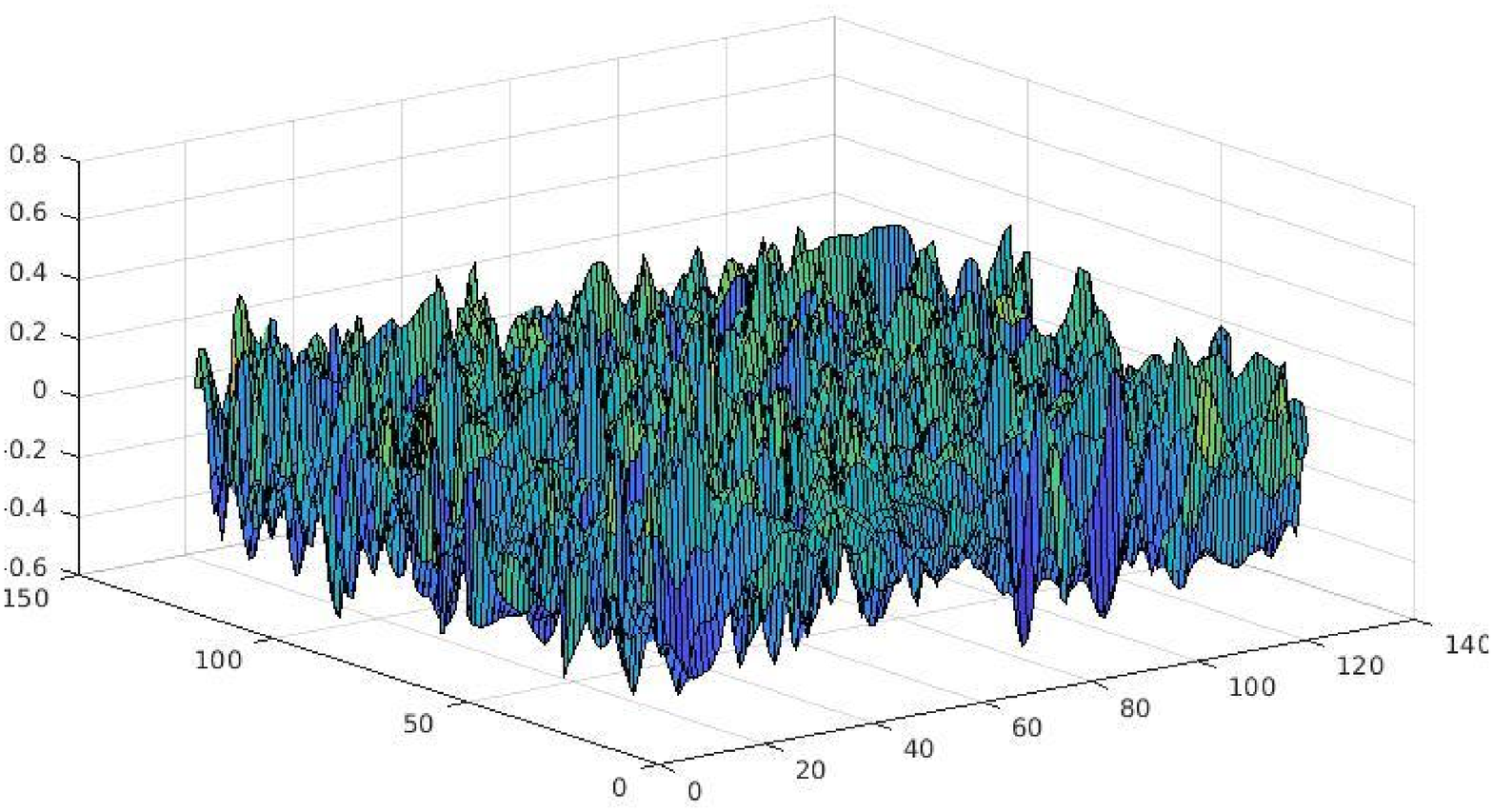}\\
c. & d. \\
\includegraphics[width=.48\textwidth]{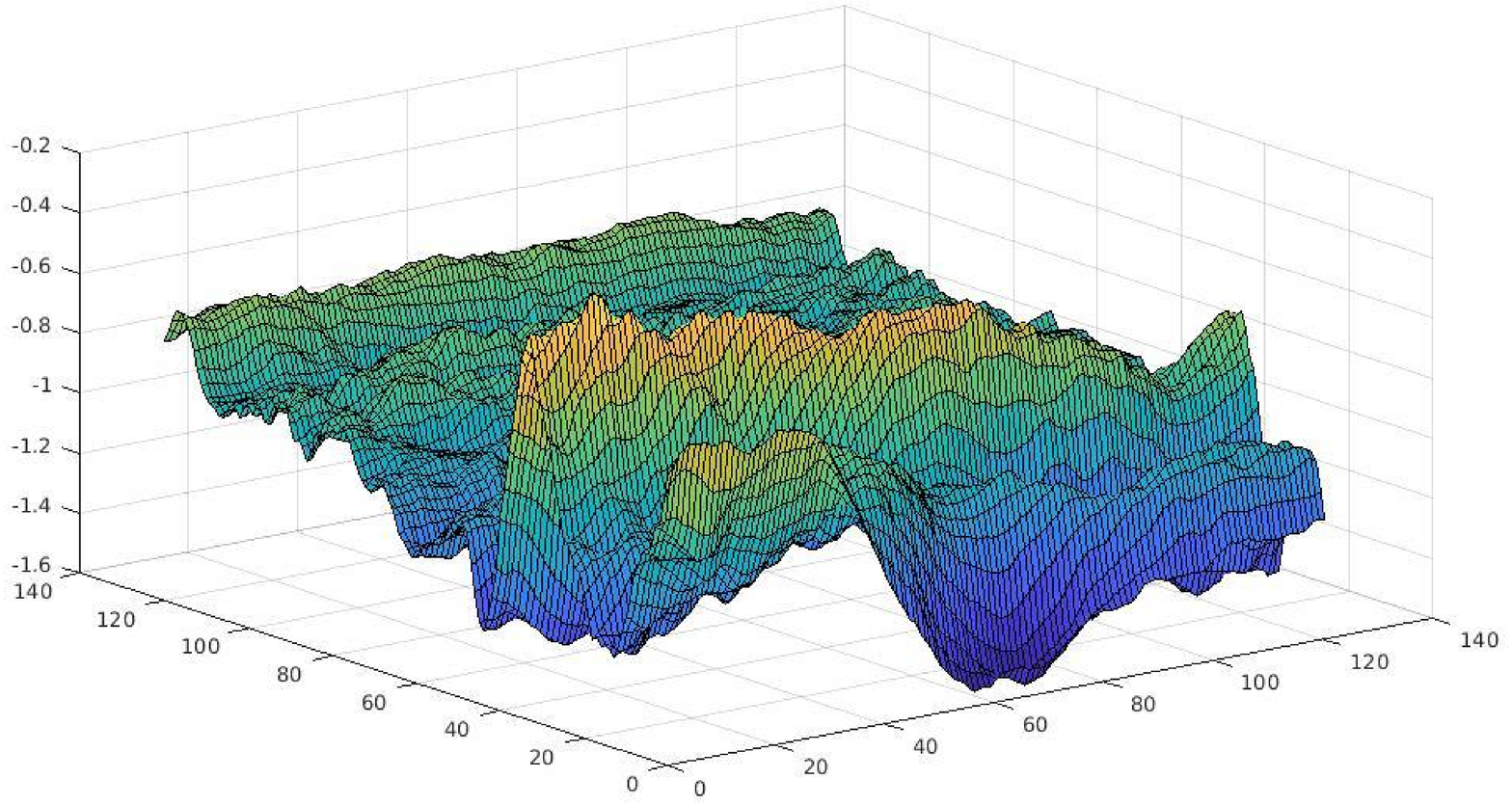}&
\includegraphics[width=.48\textwidth]{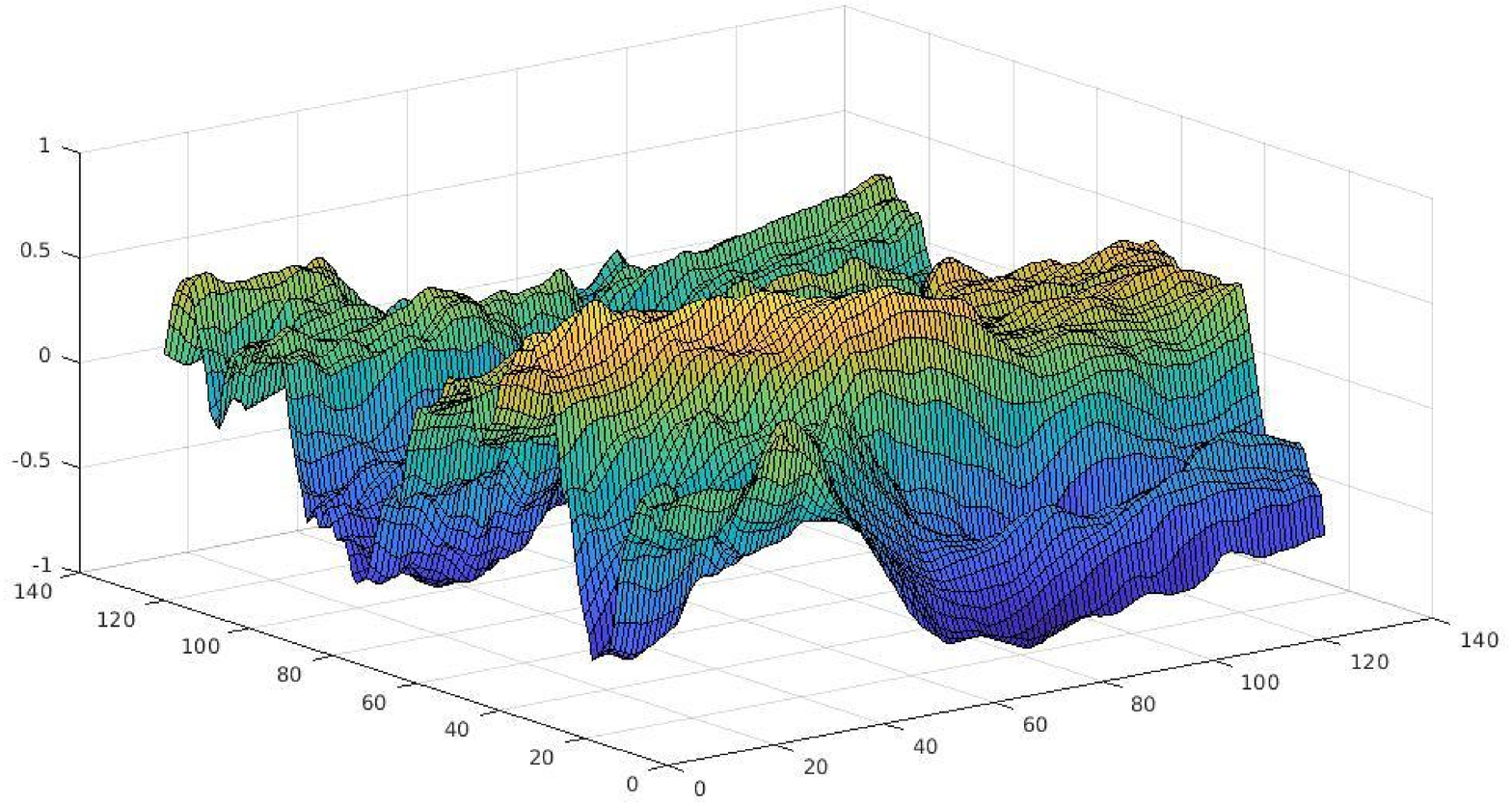}\\
e. & f. \\
\includegraphics[width=.48\textwidth]{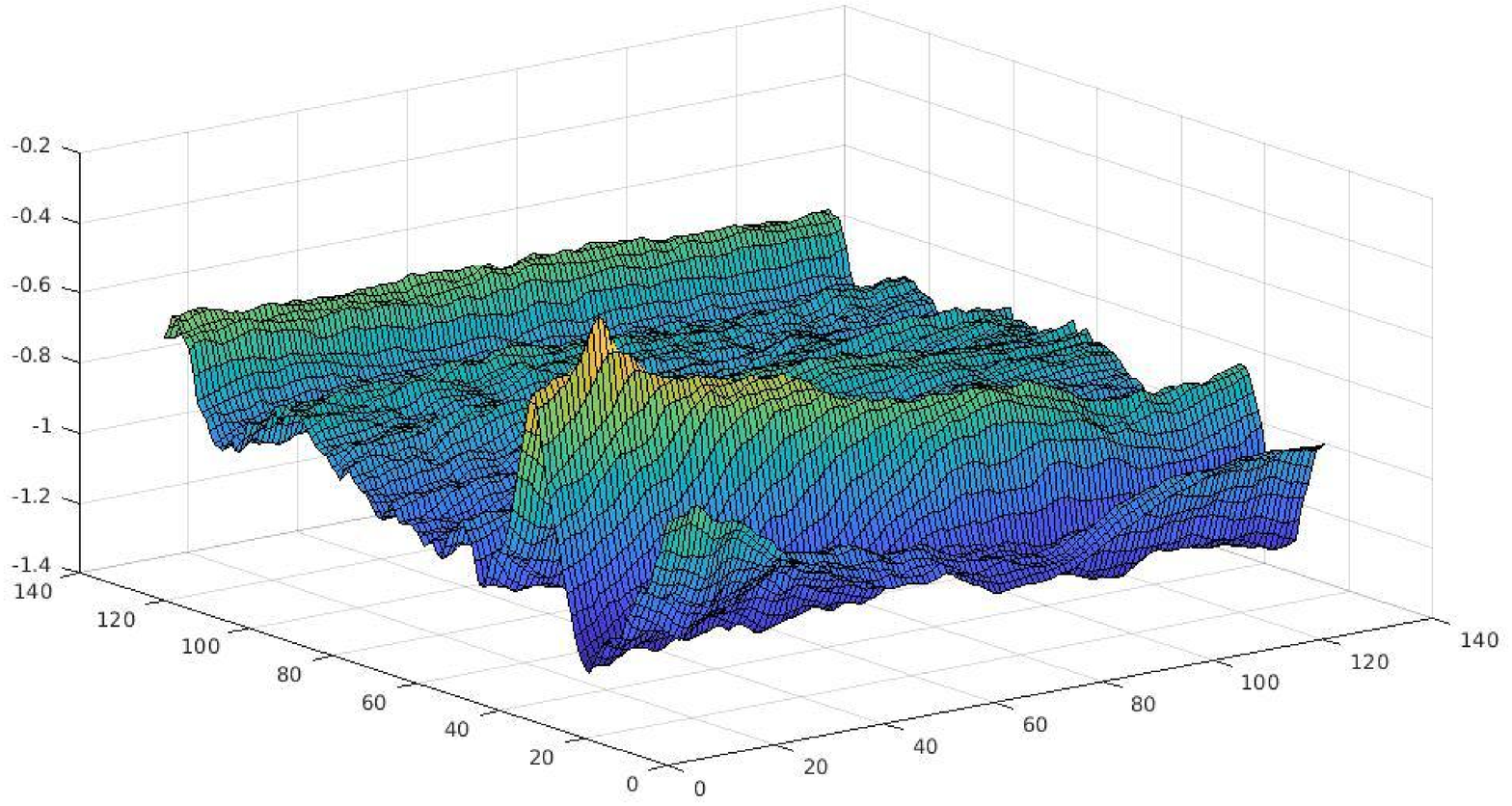}&
\includegraphics[width=.48\textwidth]{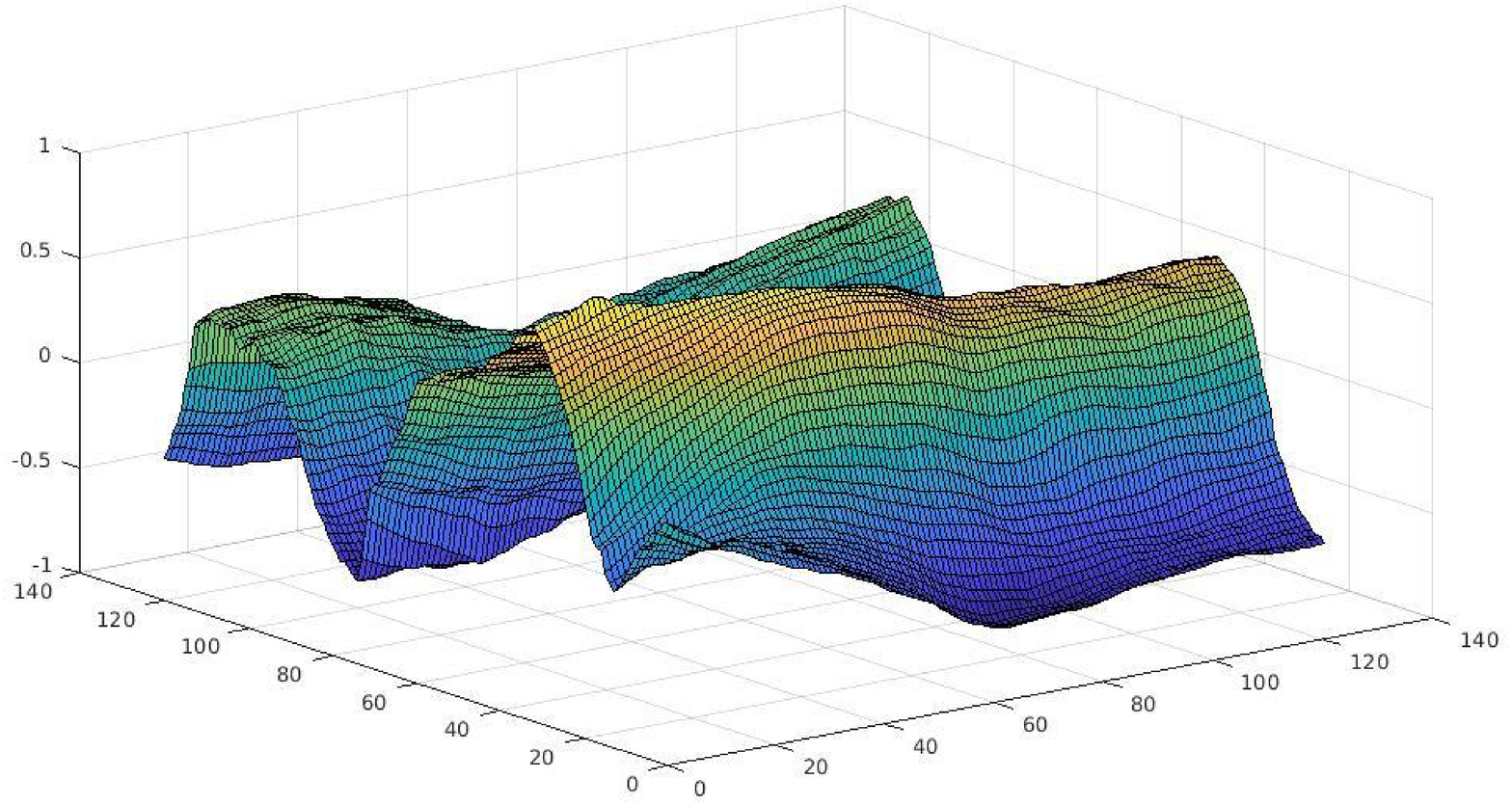}\\
g. & h. \\
\end{array}$
\caption{
\emph{a.} local ST patch, \emph{b.} corresponding image patch. 
\emph{c,e,g} Error surface for -sum of absolute difference metric for
convolution with kernel sizes of 9, 21 and 41. \emph{d,f,h} normalised zero mean correlation value surface
for kernel sizes of 9, 21, 41. 
}
\label{fig:ST_8}
\end{figure}

Figure ~\ref{fig:ST_8} compares block matching done with sum of absolute difference of ST transformed images against 
normalised zero mean correlation of the original brightens function for the same patch.
For ease of comparison - the sum of absolute difference is plotted so in both cases peaks represent better matches.

Mutlab code for a basic block matcher for a pair of ST-transformed images is.
\noindent
\begin{verbatim}
function [pts, pts2] = block_match(bx1, bx2, maxR, maxC, pflag)
% block matching function sum of absolute difference. 
%
% copyright David Sinclair 2020
%

[nr,nc] = size(bx1);
dim = 40;
br = nr-dim-2;
bc = nc-dim-2;

dog = zeros(maxR*2+1, maxC*2+1);
pts = zeros(0,2);
pts2 = zeros(0,2);
np = 0;
MR = maxR*2+1;
MC = maxC*2+1;

for r = dim+2:100:br
    r
    for c = dim+2:50:bc
        m = bx1(r-dim:r+dim, c-dim:c+dim);
        df = sum( abs(m(:)));
        if df > 1050
            dog(:) = 7000;
            for R = r-maxR:r+maxR
                i=R-r+maxR+1;
                for C = c-maxC:c+maxC
                    if( R>dim && C> dim && R<br && C<bc)
                        j=C-c+maxC+1;
                        m2 = bx2(R-dim:R+dim,C-dim:C+dim);
                        % obviously you don't have to completely evaluate this if you
                        % are in a rush:)
                        dog(i,j) = sum(abs(m(:)-m2(:)));
                    end
                end
            end
            
            
            [v,idx ] = min(dog(:));
            if v < 6000
                [rr,cc] = ind2sub([MR,MC],idx);
                np = np+1;
                pts(np,:) = [r,c];
                pts2(np,:) = [(r+rr-maxR), (c+cc-maxC)];
            end
            
            if 0 % use this to view the error surface of a local patch.
                figure(pflag +6)
                surf(dog)
                df = 9;
            end
        end        
    end
end

drcs = [pts,pts2];

if( pflag > 0 )
    cim = uint8(zeros(nr,nc,3));
    cim(:,:,1) = bx1(:,:)*60 + 38;
    cim(:,:,2) = bx2(:,:)*20 + 68;
    cim(:,:,3) = bx1(:,:)*60 + 38;
    
    figure( pflag + 10 );
    imagesc( cim )
    colormap gray
    hold on
    for i=1:np
        r = pts(i,1);
        c = pts(i,2);

        mr = pts2(i,1);
        mc = pts2(i,2);

        plot( [c;mc], [r;mr],'y' );
            
        plot( c, r,'mx' );
        plot( mc, mr,'g+' );
        
    end
    hold off 
end

return
\end{verbatim}

\subsection{Matching function stability.}

There is a trade off between the increased cost using a larger area to perform
correlation type matching and the specificity of any match. 
Figure  ~\ref{fig:ST_8} shows the error surface for various sizes of patch centred about
a point in the image pair in figure  ~\ref{fig:ST_6} for comparison standard zero mean cross
correlation for the local brightness function are included.

\section{Conclusions}
\label{section:4}
The ST-transform provides a fast and flexible basis for low level feature extraction from images.
The robust nature of the quantisation makes it a good basis for stereo matching and also for glyph
spotting or drivable region detection in applications like self driving cars or general OCR.
Matlab for some ST-transform methods is available at https://github.com/imensedave/ST-transform\_methods.
The methods used in this paper are used for a interactive highway sign reading demo at
https://www.imense.co.uk/HWAY.html

The ST-transform points out blank areas of an image that are not going to be great for matching. In general a
feature for frame to frame matching is going to need to contain a minimum turning angle in the boundary of a
ST-transform derived feature. On reflection this paper represents the last \emph{Hurrah}  of
brutish correlation based matching with future methods liable to focus exclusively
on more direct convolution net based approaches.

\bibliographystyle{splncs}
\bibliography{generalFeat,vggroup}
\end{document}